\documentclass[conference]{IEEEtran}
\usepackage{times}

\usepackage[numbers]{natbib}
\usepackage{multicol}
\usepackage[bookmarks=true,hidelinks]{hyperref}

\usepackage{amsmath, amssymb, amsfonts}
\usepackage{cuted}
\usepackage{caption}
\usepackage{listings}
\usepackage{xcolor}
\usepackage{graphicx, subcaption}
\usepackage{threeparttable}
\usepackage{multirow}
\usepackage{booktabs}
\usepackage{siunitx}
\usepackage{colortbl}
\usepackage{hhline}

\definecolor{codeblue}{RGB}{5, 75, 175}
\definecolor{codegreen}{RGB}{0, 120, 40}
\definecolor{codepurple}{RGB}{120, 40, 180}
\definecolor{lvlPerfect}{HTML}{1B9E77}
\definecolor{lvlHigh}{HTML}{1F78B4}
\definecolor{lvlMid}{HTML}{FFB000}
\definecolor{lvlLow}{HTML}{D73027}
\definecolor{lvlCaution}{HTML}{6A3D9A}

\begin{document}

\title{CRISP: Contact-Rich Robotic Simulation Platform with Extensive Geometries and Contact Solvers}

\author{\authorblockN{Somang Lee, Sunkyung Park, Jinhee Yun, Seoki An, and Dongjun Lee}
\authorblockA{Department of Mechanical Engineering, Seoul National University\\
Email: \{hopelee,sunk1136,yjhs0932,seoki97s,djlee\}@snu.ac.kr}}

\maketitle

\begin{strip}
  \centering
  \includegraphics[width=\textwidth]{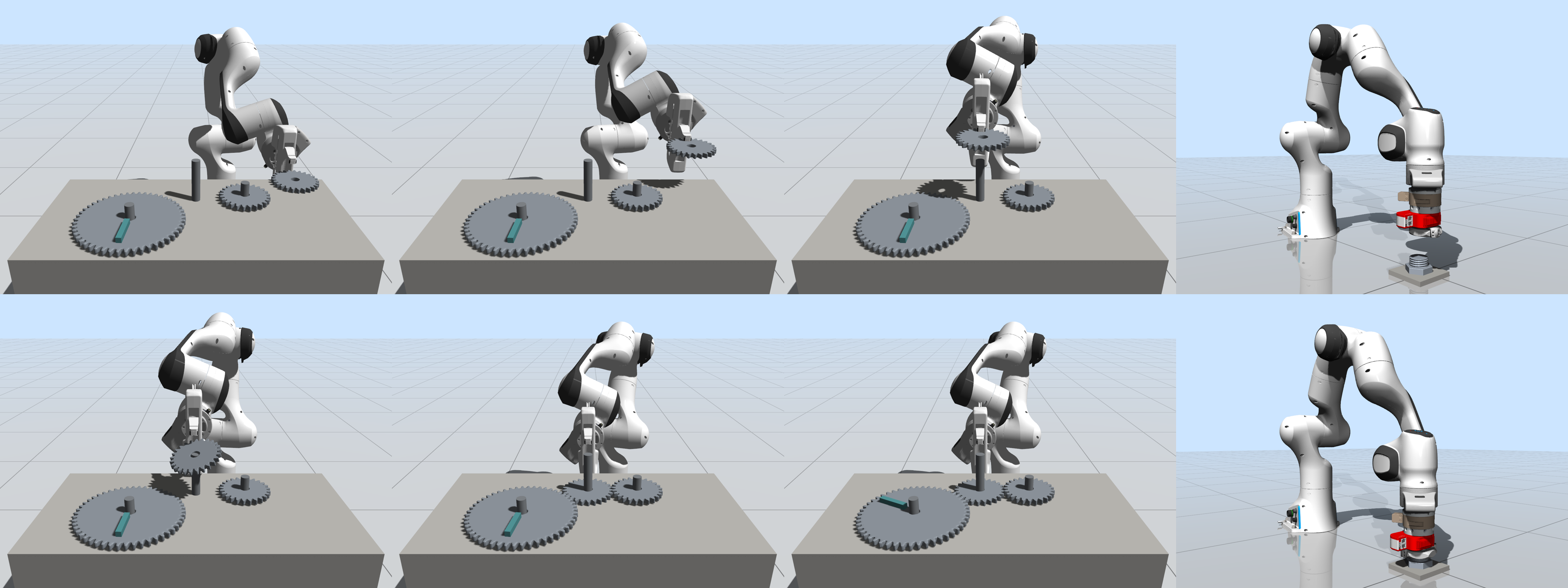}
  \captionsetup{hypcap=false}
  \captionof{figure}{
    \textbf{Snapshots of demonstration scenarios using our simulator, CRISP.}
    Left: Gear insertion and driving.
    Right: Bolt-nut assembly.
    Collision geometries are represented using meshes and SDFs for reliable collision detection, and intensive multi-contact interactions are simulated faster than real time using a robust and accurate contact solver.
  }
  \label{fig:highlight}
\end{strip}

\begin{abstract}
We present CRISP (Contact-RIch Simulation Platform), a high-fidelity physics engine tailored for complex multi-contact simulations such as tight-tolerance robotic manipulation.
Achieving high physical fidelity in robotic simulation requires both expressive modeling of geometry and contact interactions, as well as accurate numerical resolution via robust collision detection and contact solvers.
However, existing simulators often either rely on limited support for geometric representations and simplified modeling of contact interactions, or employ numerical resolution methods whose accuracy or robustness is inherently constrained.
Accordingly, we develop a new simulator that supports diverse geometric representations with accurate optimization-based collision detection, and combines contact modeling with robust augmented Lagrangian-based contact solvers.
This integration enables efficient and consistent detection of contact information across complex geometries while accurately resolving multi-contact constraints without problematic relaxations, which is essential for simulating contact-intensive and sharp interactions.
We validate the physical fidelity of our simulator against state-of-the-art platforms and further demonstrate its capabilities through complex robotic manipulation scenarios.
CRISP is publicly available at \href{https://github.com/INRoL/crisp}{https://github.com/INRoL/crisp}\footnote{Project website: \href{https://inrol.github.io/crisp/}{https://inrol.github.io/crisp/}.}.
\end{abstract}

\IEEEpeerreviewmaketitle

\section{Introduction} \label{sec:introduction}
Simulation has emerged as an indispensable component for the advancement of modern robotics, providing a safe, scalable, and cost-effective alternative to real-world experimentation.
Following the development of diverse physics engines~\cite{coumans2021bullet, hwangbo2018raisim, lee2018dart, makoviychuk2021isaac, smith2008ode, tedrake2019drake, todorov2012mujoco}, they have been widely adopted as versatile platforms that facilitate algorithm verification, system identification, mechanical design, and synthetic data acquisition~\cite{choi2021use}.
More recently, this utility has further evolved into serving as a foundational data generation platform for the advanced robotic intelligence~\cite{mittal2025isaac, nvidia2025cosmos, zakka2025mujoco}.
In this paradigm, high physical fidelity is crucial to ensure that the learned policy internalizes the underlying dynamics required for reliable zero-shot transfer to the physical world.

Since simulation is fundamentally a process of modeling physical phenomena and numerically solving the resulting models, high physical fidelity critically depends on both the expressiveness of the physical model and the accuracy of its numerical resolution.
In robotic simulation, particularly for contact-rich applications, these two aspects are largely governed by the underlying geometries and the handling of the contact interactions.

From the modeling perspective, physical fidelity is determined by the expressive scope of the simulator in representing geometry and contact interactions.
In terms of geometric representations, many widely used engines do not natively support signed distance field (SDF) representations (see Table~\ref{tab:comparison}), which provide a continuous description of geometry well suited for accurate contact modeling~\cite{macklin2020sdf}.
By contrast, Isaac Sim~\cite{makoviychuk2021isaac} supports SDF-based collision handling by processing imported meshes into voxelized distance fields, enabling scalable collision detection but limiting the use of custom analytic SDFs~\cite{omni2026sdf}, whereas MuJoCo~\cite{todorov2012mujoco} allows user-defined SDF geometries with greater flexibility while not supporting collision handling between general non-convex meshes.
Beyond geometric representations, existing simulators also substantially differ in modeling contact interactions.
Formally, the constrained dynamics with contact is naturally modeled as a nonlinear complementarity problem (NCP)~\cite{stewart1996lcp} which imposes the non-penetration and friction conditions; yet this formulation remains highly challenging to solve directly due to its non-smoothness and non-convexity~\cite{kaufman2008staggered}.
Consequently, most contemporary simulators relax the original NCP into more tractable formulations to ensure stable performance.
For instance, ODE~\cite{smith2008ode}, Bullet~\cite{coumans2021bullet}, and DART~\cite{lee2018dart} linearize the friction cone to formulate a linear complementarity problem (LCP).
While this formulation is more tractable, such approximation violates isotropy, leading to undesirable biased friction~\cite{lidec2024contact}.
Another popular appoximation is to convexify the original NCP to a cone complementarity problem (CCP)~\cite{anitescu2006ccp, todorov2014ccp} by relaxing the Signorini condition (i.e., normal complementarity), which is adopted in MuJoCo and Drake~\cite{tedrake2019drake}.
While this convexification allows the use of powerful convex optimization techniques, it induces \textit{gliding} artifacts during sliding~\cite{castro2023sap}.
On the other hand, Isaac Sim employs a position-based dynamics formulation that prioritizes numerical stability over strict physical consistency~\cite{muller2020xpbd}.

From the perspective of numerical resolution, the achievable physical fidelity of robotic simulation is further influenced by collision detection and contact constraint resolution.
In particular, the geometric validity and consistency of collision detection are inherently constrained by the supported combinations of geometric representations.
For instance, MuJoCo employs sampling-based and gradient-descent methods for SDF--SDF collision detection~\cite{mujoco2025sdf}, which are effective for smooth convex geometries but can suffer from local minima in complex configurations.
On the other hand, voxelized SDF in Isaac Sim enables robust collision detection between arbitrary shapes~\cite{macklin2020sdf}, yet the discretization may distort the original geometry and fail to capture fine-grained interactions~\cite{narang2022factory}.
Similarly, the choice of contact solvers is often constrained by the underlying contact modeling formulation.
Many simulators, including ODE, Bullet, DART, and Isaac Sim, rely on projected Gauss-Seidel (PGS)-type first-order methods, which lack convergence guarantees for ill-conditioned or tightly coupled multi-contact constraints~\cite{yoon2022assembly}.
In contrast, MuJoCo and Drake formulate contact resolution as convex optimization problems and apply second-order methods such as Newton method~\cite{castro2023sap}, enabling higher accuracy and improved convergence properties.

\begin{table}[t]
  \centering
  \begin{threeparttable}
    \begin{tabular}{l|ccc}
      \hline Physics engine
        & Geometry\tnote{1}   & Formulation & Solver     \\
      \hline \hline
      ODE~\cite{smith2008ode}
        & Mesh                & LCP         & Direct/PGS \\
      Bullet~\cite{coumans2021bullet}
        & Mesh                & LCP         & Direct/PGS \\
      DART~\cite{lee2018dart}
        & Mesh                & LCP         & Direct/PGS \\
      MuJoCo~\cite{todorov2012mujoco}
        & Mesh\tnote{3}\,/SDF & CCP         & Newton     \\
      Drake~\cite{tedrake2019drake}
        & Mesh                & CCP         & Newton     \\
      RaiSim~\cite{hwangbo2018raisim}
        & Mesh                & -\tnote{2}  & Bisection  \\
      Isaac Sim~\cite{makoviychuk2021isaac}
        & Mesh/SDF\tnote{4}   & PBD         & PGS/TGS    \\
      \hline
      CRISP (ours)
        & Mesh/SDF/DSF        & NCP         & AL-variant \\
      \hline
    \end{tabular}
    \begin{tablenotes}
      \item[1] Primitives and convexes are omitted; they are supported by all.
      \item[2] RaiSim uses a CCP variant to enforce the Signorini condition.
      \item[3] MuJoCo convexifies meshes except for mesh--SDF pairs.
      \item[4] Isaac Sim utilizes internally processed voxelized SDFs.
    \end{tablenotes}
  \end{threeparttable}
  \caption{\textbf{Comparison of existing physics engines and CRISP.}}
  \label{tab:comparison}
\end{table}

Motivated by these considerations, we present CRISP, a high-fidelity physics engine tailored for contact-rich robotic simulations.
CRISP addresses geometric and contact modeling by supporting a diverse set of geometric representations, including meshes and function-based formulations, and by directly considering the original nonlinear complementarity formulation for both rigid and compliant contacts.
These modeling capabilities are complemented by accurate optimization-based collision detection algorithms tailored to function-based geometric representations and a robust contact solver based on augmented Lagrangian (AL) theory~\cite{lee2025val}.
This integration enables the efficient and consistent detection of numerous contact points across complex geometries, while accurately resolving tightly coupled multi-contact constraints without problematic relaxations, which is essential for simulating contact-intensive and sharp interactions.
Our contributions are summarized as follows:
\begin{itemize}
  \item To our knowledge, this is the first physics engine both capable of accurately capturing complex collision geometries and resolving multi-contact interactions without unphysical relaxations, by providing various optimization based collision detection algorithms and contact solvers that are highly accurate and efficient compared to the other approaches in existing simulators.
  \item Our framework is validated by complex manipulation demonstrations and comparative evaluation against state-of-the-art platforms, exhibiting superior physical fidelity.
  \item CRISP is publicly available, with documentation, examples, and videos provided on the project website.
\end{itemize}

The rest of the paper is organized as follows.
In Sec.~\ref{sec:related_works}, we review typical contact handling methods in existing physics engines.
Sec.~\ref{sec:simulator} provides a high-level overview of CRISP, with Sec.~\ref{sec:collision_detection} introducing its diverse geometric representations and Sec.~\ref{sec:constraint_solving} describing its AL-based contact solvers.
Various comparative evaluations and demonstrations are presented in Sec.~\ref{sec:evaluation}, followed by concluding discussions in Sec.~\ref{sec:conclusion}.

\begin{figure*}[t]
    \centering
    \includegraphics[width=\textwidth]{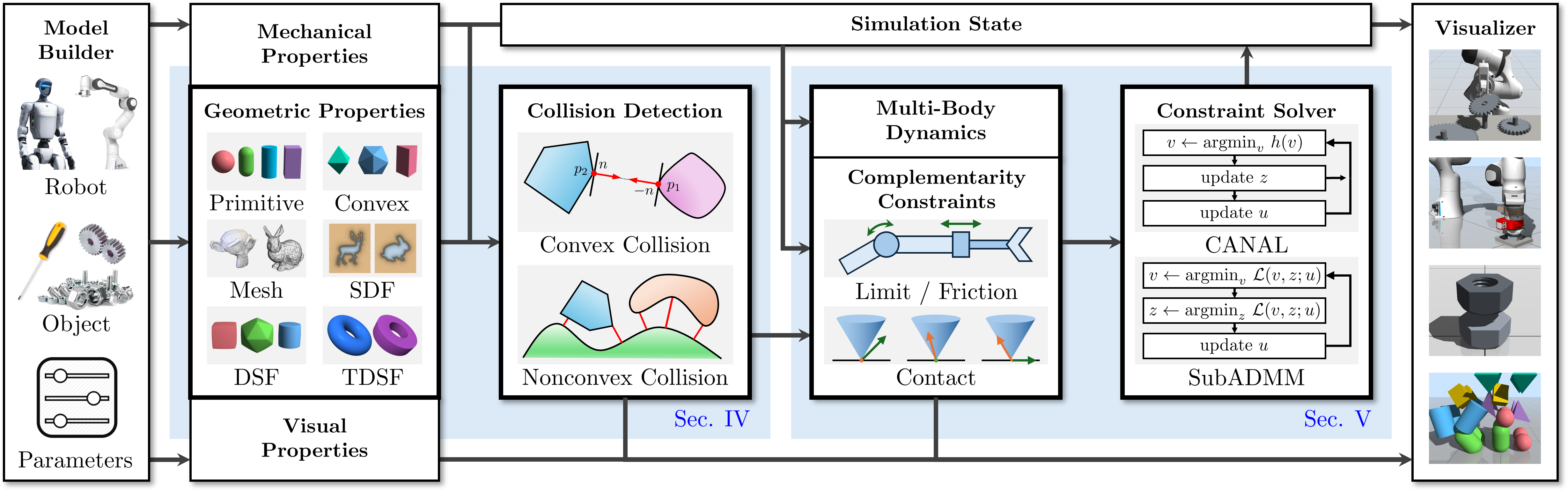}
    \caption{
      \textbf{High-level architecture of CRISP.}
      The model builder defines the low-level structure \texttt{model\_t}, from which \texttt{data\_t} is instantiated.
      At each step, collision detection and constraint solving are performed, and the resulting state is visualized.
    }
    \label{fig:overview}
\end{figure*}

\section{Related Works} \label{sec:related_works}
Contact interactions are typically managed through the interplay between collision detection and contact solving.
Table~\ref{tab:comparison} summarizes the methodological choices of widely-used physics engines in robotics, alongside the proposed CRISP.

\subsection{Collision Detection}
Collision detection serves to identify the contact informations between geometries and construct the contact constraints.
Among various geometric representations, primitive shapes (e.g., spheres, capsules, boxes) and convex polytopes are the most simulation-friendly form, as they enable explicit solutions or efficient algorithms for collision queries~\cite{gilbert2002gjk, montaut2024gjk++, snethen2008mpr}.
In contrast, while triangle meshes are the \textit{de facto} standard for graphical purpose, their discrete and non-convex nature poses significant challenges for robust collision detection~\cite{wang2021ccd}.
Consequently, most physics engines either offer restricted support (e.g., Drake) or rely on computationally expensive detection algorithms~\cite{pan2012fcl} for mesh collisions; thus strongly recommending convex decomposition~\cite{mamou2009vhacd, wei2022coacd} as a preprocessing step.

From a simulation perspective, function-based representations offer several advantages over discrete meshes.
Most notably, SDFs enable efficient and robust collision detection even in complex configurations~\cite{macklin2020sdf}.
However, the applicability of SDFs is often limited by the availability of analytic distance functions, and the efficient storage and querying of such representations remain challenging~\cite{frisken2000sdf}.
Another promising form is the differentiable support function (DSF) to represent smooth convex shapes, which is independently proposed in \cite{halm2023addressing} and \cite{lee2023dsf}.
This representation provides second-order surface information with respect to the supporting direction, enabling fast and robust collision detection between smooth convex shapes~\cite{an2024dsf}.

\subsection{Contact Solver}
Since multi-contact NCP originating from rigid body dynamics with frictional contacts is inherently difficult to solve directly due to its non-smooth and non-convex nature~\cite{kaufman2008staggered}, most contact solvers address relaxed formulations.
The LCP formulation, which linearizes the friction cone into a pyramid, allows the use of direct methods such as Lemke's~\cite{lloyd2005lemke} and Dantzig's algorithms, or per-contact iterative methods like PGS algorithm~\cite{horak2019similarities}, which is widely adopted in game and graphics communities~\cite{andrews2022contact} as well as in several robotic simulators including ODE, Bullet, and DART, despite its poor convergence properties and the introduction of anisotropic friction~\cite{lidec2024contact}.
Another widely used relaxation is the CCP formulation, whose convexification enables the used of powerful convex optimization techniques, such as Newton method, as adopted in MuJoCo and Drake.
To mitigate the \textit{gliding} artifacts associated with CCP, Raisim~\cite{hwangbo2018raisim} adopts an alternative formulation, yet it introduces incorrect friction direction~\cite{lidec2024contact}.

To avoid these relaxation-induced artifacts, there have been several attempts to address the original NCP.
For instance, \citet{howell2022dojo} adopted an interior point method and \citet{macklin2019ncp} utilized a complementarity function, though both approaches eventually induce numerical smoothing.
\citet{macklin2019tgs} also proposed a variant of the PGS method later employed in Isaac Sim, which relies on the position-level dynamics formulation.
More recently, AL methods have been explored to gradually address the multi-contact NCP by solving a sequence of relaxed subproblems.
\citet{carpentier2024simple} introduced an ADMM-based solver using a dual formulation, while \citet{lee2025val} proposed two variations of the AL method using a primal-dual formulation.

\section{CRISP: Contact-Rich Simulation Platform} \label{sec:simulator}

\subsection{System Overview}
We develop CRISP to enable efficient and accurate simulation of contact-rich robotic systems across a wide range of manipulation scenarios.
The overall architecture of CRISP is inspired by MuJoCo and structured into four primary components, as depicted in Fig.~\ref{fig:overview}:

\begin{itemize}
  \item The initial stage involves \textbf{model building}, where the user defines the simulation scenario by specifying the mechanical properties of bodies and joints, geometric properties attached to each body, and various solver parameters to tune the simulation fidelity.

  \item The \textbf{collision detection} engine identifies pairs of geometries where a collision is imminent or has already occured under the current simulation state, extracting contact information such as penetration depths, contact points and normals.

  \item The \textbf{constraint solver} is a core component of the CRISP physics engine and is responsible for solving multi-body dynamics subject to contact and joint constraints; in particular, CRISP provides two tailored AL-based solvers, CANAL and SubADMM, ensuring both computational efficiency and physical accuracy.

  \item The \textbf{visualization} is handled by a built-in OpenGL-based forward renderer, integrated directly into the simulator to provide lightweight yet efficient visual feedback.
\end{itemize}

Our framework is implemented in C++20, leveraging the Eigen library~\cite{eigen2010eigen} for high-performance linear algebra with SIMD vectorization, and adopts a custom arena-based allocation scheme to minimize dynamic memory allocation during simulation steps, mitigating allocation overhead and memory fragmentation while ensuring predictable and stable performance.

\subsection{Model Building}
The user can configure custom simulation scenarios consisting of various single or articulated bodies.
Currently, this is achieved through a C++ API that allows for the programmatic building of a hieracical model structure.
This tree-like high-level description reflects the kinematic connectivity of the robot and its environment, where each node contains the following essential information:

\begin{itemize}
  \item \textbf{Mechanical properties} of each body and joint are defined to specify inertial characteristics, including mass and inertia, as well as joint configurations such as joint types, axes of motion, mechanical limits and frictions.

  \item \textbf{Geometric properties} define the visual or collision shapes associated with each body through diverse geometric representations, each assigned with surface attributes such as friction coefficient and compliance.

  \item \textbf{Visual properties} are assigned to enhance the quality of the visual output, managing attributes for the built-in renderer such as material colors, textures, and lighting.

  \item \textbf{Simulation options} consist of the parameters to tune the solver's performance and fidelity, including time step size, solver type, and convergence criteria.
\end{itemize}

To illustrate the ease of model definition, the following code snippet demonstrates the construction of a simulation environment comprising a robotic arm defined via a URDF file and a single box-shaped object.
\begin{lstlisting}
auto builder = crisp::make_model_builder();
builder->addRobot({.path = "franka.urdf"});
auto& body = builder->addBody(
  {.pos = {0, 1, 1}, .mass = 10,
   .inertia = {1, 1, 1, 0, 0, 0}});
auto& geom = body.addGeom(
  {.type = crisp::geometry_e::box,
   .size = crisp::make_box_size(1, 1, 1)});
auto model = builder->build();
\end{lstlisting}
While certain parameters are mandatory to construct the valid simulation model, others are optional and provided with sensible default values.

Once the high-level model description is complete, it is compiled into a compact low-level data structure \texttt{model\_t}, which organizes kinematic hierarchies into a flattened memory layout to maximize cache efficiency and optimize simulation performance.
This static blueprint, which remains constant throughout the simulation, is then used to instantiate \texttt{data\_t} as a mutable workspace that stores all time-varying quantities including generalized positions, velocities, contact informations, and solver states.
This decoupled architecture, similar to that of MuJoCo, facilitates the multi-threaded execution of multiple simulations in parallel while sharing a single \texttt{model\_t} instance, which aids in sampling-based control and high-fidelity dataset generation for robotic intelligence.

\section{Collision Detection} \label{sec:collision_detection}
The collision detection module identifies contacts between geometries and provides the contact information required for the constraint solver.
This section describes the supported geometric representations and the key collision detection algorithms implemented in CRISP.

\subsection{Geometry Representations}
CRISP supports a diverse range of geometric representations, allowing users to select the most appropriate one based on their specific requirements for computational efficiency and geometric fidelity.
These include primitives (planes, spheres, capsules, boxes, and cylinders), convex hulls, and meshes, as well as function-based representations such as SDFs, DSFs, and our novel TDSFs.
Each function-based representation is discussed in detail below.

\subsubsection{SDF}
An SDF defines the geometry through a scalar function which returns the signed distance from a point to the closest point on the surface of the shape.
While SDFs offer the advantage of representing arbitrary non-convex shapes with a single geometric representation, they inherently increase the complexity of the collision detection problem.
CRISP provides a flexible interface for defining SDF geometries, allowing users to register custom SDFs through callable functions that can be further parameterized for shape variations.
To register a custom SDF, users need only provide a callable function that evaluates the signed-distance value at a given local query point.
An analytic gradient may optionally be supplied; otherwise, CRISP approximates it using finite differences.

\subsubsection{DSF}
\begin{figure}[t]
  \centering
  \begin{subfigure}{0.9\columnwidth}
    \centering
    \begin{subfigure}{0.499\columnwidth}
      \includegraphics[width=\columnwidth]{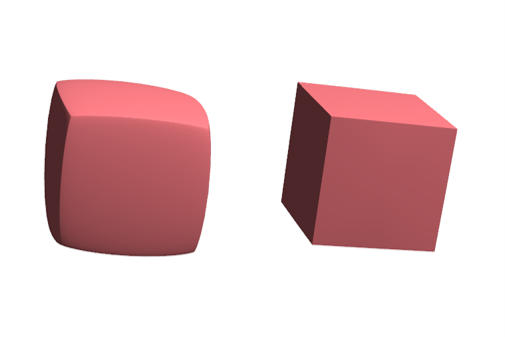}
      \caption{Cube}
    \end{subfigure}\hfill
    \begin{subfigure}{0.499\columnwidth}
      \includegraphics[width=\columnwidth]{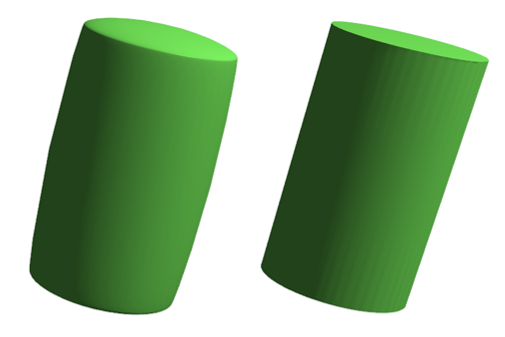}
      \caption{Cylinder}
    \end{subfigure}\hfill
  \end{subfigure}
  \begin{subfigure}{0.9\columnwidth}
    \centering
    \includegraphics[width=\columnwidth]{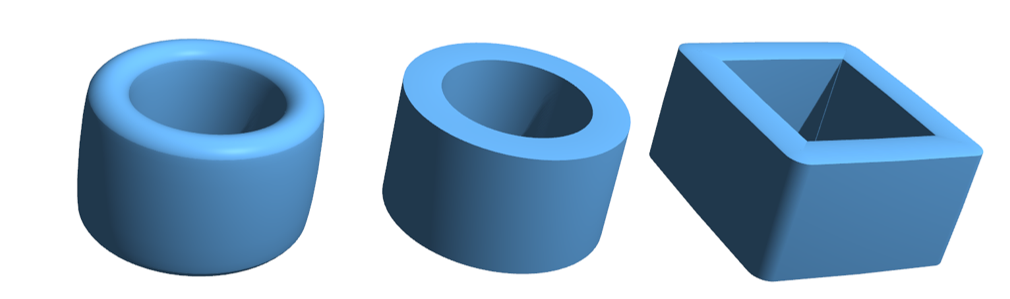}
    \caption{Torus}
  \end{subfigure}
  \caption{
    \textbf{Visualizations of DSF and TDSF geometries.}
    It is possible to specify the degree of approximation by choosing the smoothness parameter, as described in Appendix~\ref{app:dsf}.
  }
  \label{fig:dsf}
\end{figure}

A support function $h:\mathbb{R}^3\to\mathbb{R}$ for a convex set $\mathcal{C}\subset\mathbb{R}^3$ is defined by
\begin{equation*}
  h(x) = \sup_{s\in\mathcal{C}} x \cdot s
\end{equation*}
where $\cdot$ denotes the standard inner product.
For closed convex sets, there exists a one-to-one correspondence between the convex set and its support function~\cite{rockafellar1970convex}, allowing $\mathcal{C}$ to be uniquely defined by $h$.
Building upon this, \citet{lee2023dsf} introduced the DSF, which employs twice-differentiable support functions to represent strictly convex smooth shapes, which enables collision detection to be formulated as a differentiable optimization problem.

DSF offers several advantages over simple primitive shapes and convex hulls, as it formulates the collision detection problem as strongly convvex, guaranteering a unique contact point and normal which is further differentiable with respect to the geometry pose.
From a simulation perspective, this ensures temporally consistent contact generation, making DSF particularly suitable for delicate manipulation between smooth convex shapes.
Similar to SDF, DSF refers to a general class of shape functions, with CRISP currently supporting two specific DSF implementations: a vertex-based DSF to approximate a convex hull and an axis-based DSF to approximate a solid of revolution, as illustrated in Fig.~\ref{fig:dsf}.
See Appendix~\ref{app:dsf} for detailed description of these classes.

\subsubsection{TDSF}
To further expand the range of shapes representable by DSFs, we introduce the toroidal DSF (TDSF), which defines a class of toroidal shapes by revolving a planar DSF along another planar DSF path, as illustrated in Fig.~\ref{fig:dsf}.
Specifically, given two planar DSFs $h_1,h_2:\mathbb{R}^2\to\mathbb{R}$, a TDSF $h_T:\mathbb{R}^2\times\mathbb{R}^2\to\mathbb{R}$ is defined by
\begin{equation*}
  h_T(u,v) = v_1h_1(u) + h_2(v)
\end{equation*}
where $v=(v_1,v_2)\in\mathbb{R}^2$.
Although the representable shapes by TDSFs are limited to those with convex cross-sections, they can effectively model a variety of common geometries with a hole such as tori, and tubes while maintaining the computational efficiency and robustness advantages of DSF-based collision detection.

\subsection{Collision Detection Algorithms}
Following the standard practice in robotics simulation, CRISP first performs broad-phase filtering to prune non-colliding pairs, including axis-aligned bounding box checks, after which the remaining candidates are passed to the narrow-phase for precise collision detection.
During the narrow-phase, specialized algorithms are dispatched for each geometry pair according to the collision function table.
As the efficiency and accuracy of collision detection algorithms vary, users may select the most appropriate geometry combination based on their specific requirements.
Since some of the algorithms are standard or straightforward, we focus on describing algorithms associated with function-based representations and provide the complete collision function table in Appendix~\ref{app:collision_table}.

\subsubsection{Mesh--SDF}
Following \cite{macklin2020sdf}, CRISP identifies candidate mesh triangles that overlap with the SDF geometry and applies the Frank--Wolfe algorithm to selected candidates.
This procedure is naturally parallelizable and identifies multiple contact points between two possibly non-convex shapes, ensuring robust collision detection even in complex configurations.
For high-resolution meshes, this method may generate a large number of candidate contacts, which can be numerically challenging for the constraint solver; thus CRISP retains a bounded subset, favoring spatial distribution when applicable.

\subsubsection{SDF--SDF}
For SDF pairs, we adopt the same objective used in MuJoCo~\cite{mujoco2025sdf}: solve $\min_x\phi_1+\phi_2+|\max(\phi_1,\phi_2)|$ via multi-start gradient descent over the overlapping region.
While this algorithm is applicable to arbitrary SDF pairs, it may fail to identify all contact points when the number of sampled initial points is insufficient.

\subsubsection{DSF--DSF}
For DSF pairs, we adopt the dedicated algorithm proposed in \cite{an2024dsf}.
This algorithm utilizes Riemannian optimization to ensure superlinear convergence and theoretically guaranteed robustness, which is validated to exhibit superior speed and accuracy compared to standard convex-convex collision detection algorithm.

\subsubsection{SDF--DSF}
For SDF--DSF pairs, we exploit the fact that the surface unit normal of an SDF can be readily obtained from its gradient, allowing the SDF surface to be treated analogously to a DSF within the collision optimization.
Based on this observation, we extend the DSF--DSF collision detection algorithm to SDF--DSF pairs, similarly to the extension used for DSF--TDSF collision detection.

\subsubsection{(T)DSF--TDSF}
For pairs involving TDSFs, we extend the DSF--DSF collision detection algorithm~\cite{an2024dsf} by leveraging the special structure of TDSF.
The extended algorithms also employ Riemannian optimization, ensuring accurate and efficient collision detection for both DSF--TDSF and TDSF--TDSF pairs.
The DSF--TDSF algorithm is particularly advantageous over SDF-based methods for simulating peg-in-hole scenarios, as it provides consistent contact points and normals essential for physically accurate simulation.

\section{Constraint Solving} \label{sec:constraint_solving}
Based on the contact information obtained from collision detection, the solver addresses the constrained dynamics of the system to compute the next state.
CRISP currently provides two augmented-Lagrangian-based constraint solvers, CANAL and SubADMM.
This section provides a concise description of these solvers; further algorithmic details are available in Appendix~\ref{app:constraint_solving} and \cite{lee2025val}.

\subsection{Problem Formulation}
We consider following discretized constrained dynamics:
\begin{equation} \label{eq:dynamics}
  Av = b + J^T\lambda \quad
  \text{s.t. } (Jv,\lambda)\in\mathcal{S}_c
\end{equation}
where $A\in\mathbb{R}^{n_v\times n_v}$ and $b\in\mathbb{R}^{n_v}$ represents the multi-body system dynamics~\cite{featherstone2008rbda}, $v\in\mathbb{R}^{n_v}$ is the generalized representative velocity, $J\in\mathbb{R}^{n_c\times n_v}$ is the constraint Jacobian, and $\lambda\in\mathbb{R}^{n_c}$ is the constraint impulse.
The set $\mathcal{S}_c$ represents the collective constraints of the system, which are categorized as joint limits, joint frictions, contact constraints (either rigid or compliant).
All of these constraints can be formulated as complementarity conditions, as detailed in Appendix~\ref{app:constraint_solving}, and are handled directly without numerical relaxations.

The AL-based solvers consider a decoupled reformulation of \eqref{eq:dynamics} by introducing an auxiliary variable $z\in\mathbb{R}^{n_c}$ as follows:
\begin{equation*}
  Av = b + J^T\lambda \quad
  \text{s.t. } (z,\lambda)\in\mathcal{S}_c, ~Jv = z
\end{equation*}
with the corresponding AL defined as:
\begin{equation} \label{eq:augmented_lagrangian}
  \mathcal{L} =
  \frac{1}{2}v^TAv - b^Tv + g(z) + u^T(Jv-z) + \frac{\beta}{2}\|Jv-z\|^2
\end{equation}
where $u\in\mathbb{R}^{n_c}$ is the Lagrange multiplier, and $\beta>0$ is the penalty weight, and $g$ serves to enforce $(z,\lambda)\in\mathcal{S}_c$.
At each iteration, the AL-based solver addresses the relaxed surrogate problem for the optimality of \eqref{eq:augmented_lagrangian}, and subsequently updates the multiplier $u$, gradually converging to the solution of the original problem \eqref{eq:dynamics}.
Since this AL approach maintains the feasibility of the surrogate problem, it enables stable and robust handling of complex complementarity constraints, particularly challenging rigid contacts, with least constraint violation even in poorly conditioned problems~\cite{dai2023al}.

\subsection{CANAL Solver}
The aforementioned surrogate problem in AL can be expressed as the following semismooth equation:
\begin{equation} \label{eq:canal_residual}
  r(v) = Av - b - J^T\Pi(-\beta Jv - u - \beta e)
\end{equation}
where $\Pi$ denotes the projection onto $\mathcal{S}_c$.
CANAL addresses \eqref{eq:canal_residual} via a cascaded Newton method, exploiting the integrability of the residual \eqref{eq:canal_residual} into a strongly convex function by treating the De Saxc\'e corrections as constants borrowed from the previous AL iteration.
This cascaded Newton structure enables CANAL to converge to a highly accurate solution of multi-contact NCP.
We also note that a single AL iteration in CANAL corresponds to solving a convex compliant-contact subproblem with a Newton-type method, closely related to the convex contact formulations used in MuJoCo and Drake; thus CANAL can be viewed as an AL extension that iteratively refines such relaxations~\cite{lee2025val}.

\subsection{SubADMM Solver}
While CANAL effectively resolves the multi-contact NCP, its reliance on Newton step necessitates the factorization of the Hessian matrix, which becomes computationally expensive for high degree-of-freedom (DoF) multi-body systems.
As a scalable alternative, the SubADMM solver adopts an ADMM approach~\cite{boyd2004convex} based on the decomposition of the dynamics \eqref{eq:dynamics} into subsystems (i.e., ground-rooted subtrees), each reformulated as follows:
\begin{equation*}
  A_jv_j = b_j + \sum_{i\in\mathcal{I}_j} J_{i,j}^T\lambda_i, \quad
  \text{s.t. } \Bigl(\sum_{j\in\mathcal{J}_i} J_{i,j}v_j,\lambda_i\Bigr)
  \in\mathcal{S}_{c,i}
\end{equation*}
where $\mathcal{I}_j\subset\{1,\cdots,n_c\}$ and $\mathcal{J}_i\subset\{1,\cdots,n_v\}$ are the index sets associated with the $j$-th subsystem and the $i$-th constraint, respectively, with $|\mathcal{J}_i|$ typically being 1 or 2.

Then the ADMM iteration process consists of performing an alternating update for each $v_j$ and $z_{i,j}$ from the surrogate problem and updating the multiplier $u_{i,j}$, with each step naturally parallelizable as it is decoupled across subsystems or constraints.
This structure allows SubADMM to quickly solve the high-DoF multi-contact NCP to a reasonable degree, yet CANAL yields superior accuracy via its second-order nature.
Note that while \citet{carpentier2024simple} also proposed an ADMM-based NCP solver, their approach relies on a dual formulation to introduce slack variables for impulses, which precludes the subsystem-based variable splitting utilized in SubADMM.

\section{Evaluation and Demonstration} \label{sec:evaluation}
In this section, we present a comparative evaluation of CRISP against state-of-the-art simulators, MuJoCo and Isaac Sim, followed by demonstrations that highlight its simulation capabilities in complex robotic manipulation tasks.
To quantitatively assess the physical accuracy of CRISP, we simulate several representative multi-contact scenarios and compare the results across simulators.
For each comparative evaluation, we match geometric configurations, physical parameters, and time step sizes as closely as possible across simulators; additional setup details are provided in Appendix~\ref{app:evaluation_setup}.
The experiments were conducted using MuJoCo v3.4.0 and Isaac Sim v4.5.0 on an AMD Ryzen 7 9800X3D CPU at 4.70 GHz.

\subsection{Peg Insertion} \label{subsec:peg_insertion}
\begin{figure}[t]
  \centering
  \begin{minipage}{0.49\linewidth}
    \centering
    \includegraphics[width=\linewidth]{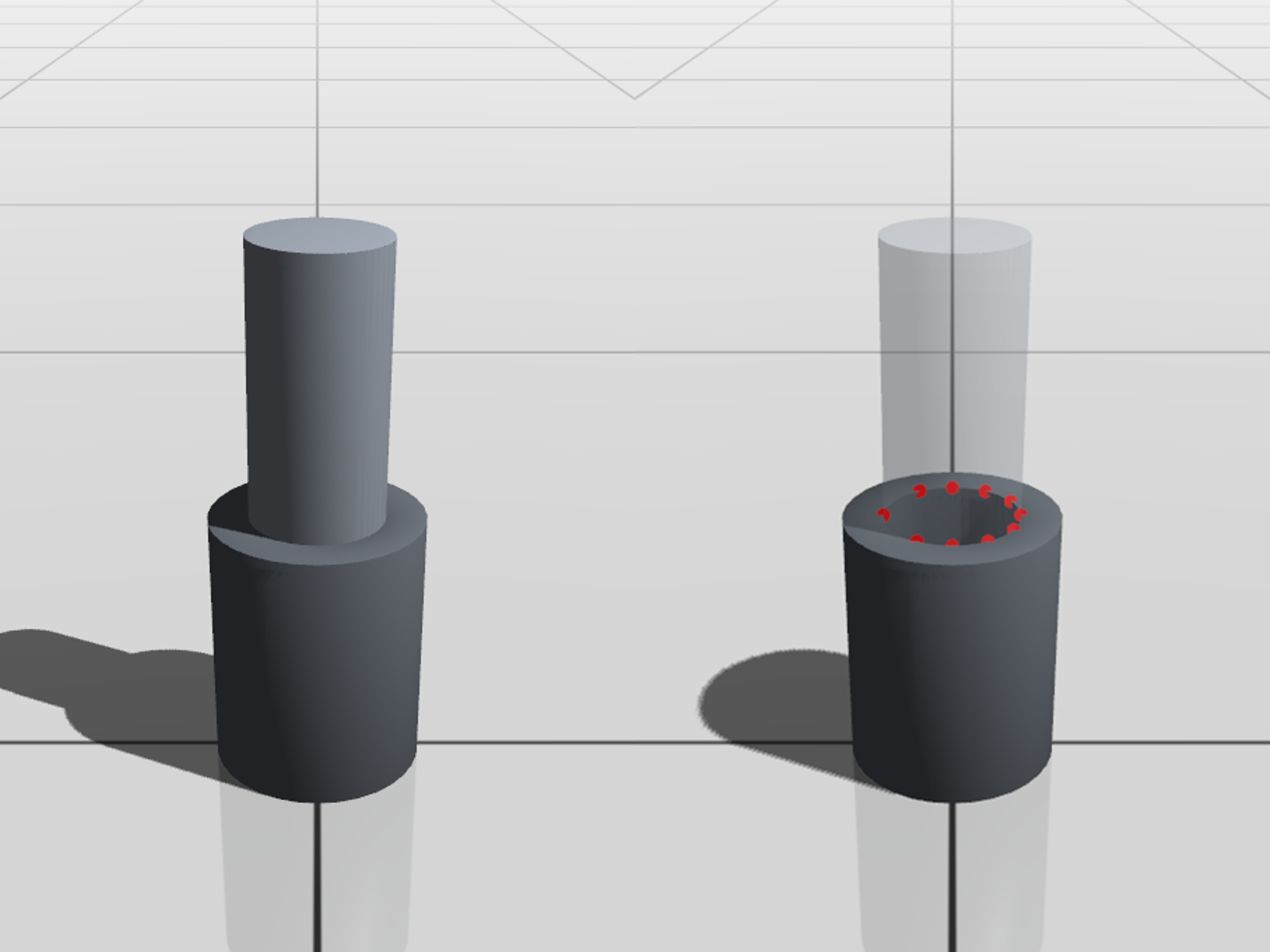}
  \end{minipage}\hfill
  \begin{minipage}{0.49\linewidth}
    \centering
    \includegraphics[width=\linewidth]{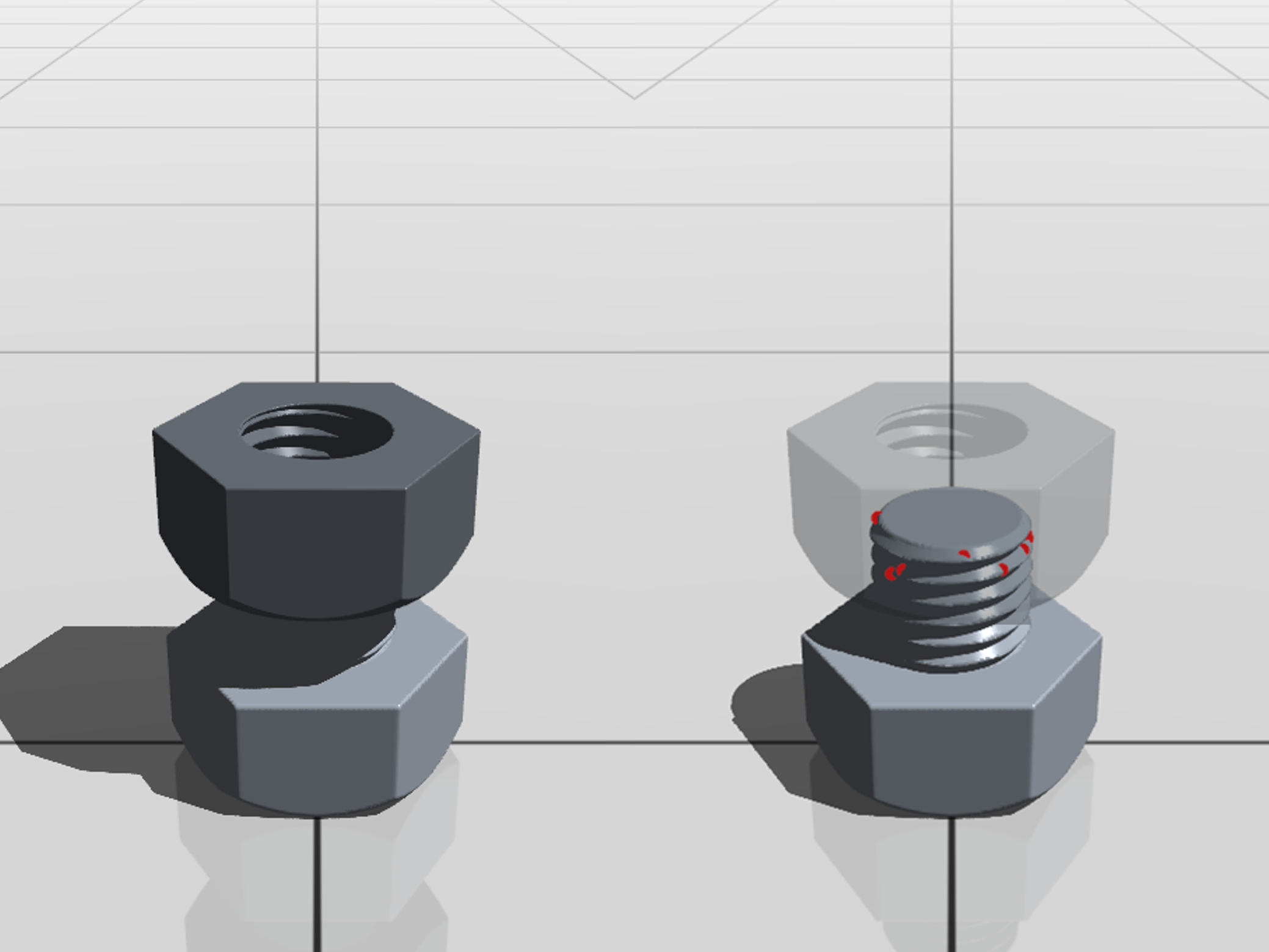}
  \end{minipage}
  \caption{
    \textbf{Visualization of assembly scenarios.}
    Initial configurations and examples of detected contact points.
    Left: peg insertion scenario.
    Right: bolt-nut assembly scenario.
  }
  \label{fig:assembly}
\end{figure}

\begin{table}[t]
  \centering
  \begin{tabular}{c|c|c|c|c|c|c}
    \hline
    \multicolumn{7}{l}{\textbf{Peg Insertion} (Sec. \ref{subsec:peg_insertion})} \\
    \hline
    \multirow{2}{*}{\shortstack{Tol. \\ (\unit{\micro\meter})}}
    & \multicolumn{4}{c|}{CRISP (CANAL/SubADMM)} & MuJoCo & Isaac Sim \\
    \cline{2-7}
    & \multicolumn{2}{c|}{DSF--TDSF} & \multicolumn{2}{c|}{SDF--SDF} & SDF--SDF & Mesh--SDF \\
    \hline
    200 & \textbf{43.649} & 80.565 & 147.61 & 143.89 & 756.31 & 47.701          \\
    100 & 25.646          & 124.46 & 107.95 & 90.769 & 1349.9 & \textbf{22.551} \\
    50  & \textbf{37.844} & 43.503 & 79.334 & 106.09 & 1328.6 & -               \\
    \hline \hline

    \multicolumn{7}{l}{\textbf{Bolt-Nut Assembly} (Sec. \ref{subsec:bolt_nut})} \\
    \hline
    \multirow{2}{*}{\shortstack{Tol. \\ (\unit{\micro\meter})}}
    & \multicolumn{4}{c|}{CRISP (CANAL/SubADMM)} & MuJoCo & Isaac Sim \\
    \cline{2-7}
    & \multicolumn{2}{c|}{Mesh--SDF} & \multicolumn{2}{c|}{SDF--SDF} & SDF--SDF & Mesh--SDF \\
    \hline
    2000 & \textbf{9.5955} & 991.59 & 61.338          & 284.18 & 65.517 & 64.549 \\
    1000 & \textbf{9.1545} & 898.42 & 46.783          & 254.96 & 68.306 & 60.018 \\
    500  & -               & 951.34 & \textbf{47.825} & 247.78 & 67.228 & -      \\
    \hline
  \end{tabular}
  \caption{
    \textbf{Average penetration depth across assembly scenarios.}
    Units are in \unit{\micro\meter}, and ommitted entries indicate failed assemblies.
  }
  \label{tab:penetration}
\end{table}

The first evaluation scenario is a peg insertion task, where a cylindrical peg (radius 2.5~\unit{\centi\meter}, height 10~\unit{\centi\meter}) is inserted into a hole with a tight tolerance of 50--200~\unit{\micro\meter}, as illustrated in Fig.~\ref{fig:assembly}.
Although MuJoCo supports mesh--SDF collision, we report its analytic SDF--SDF configuration to match the SDF--SDF baseline in CRISP and to keep the comparison tied to the same analytic geometry.
For Isaac Sim, we import the geometries as high-resolution meshes (on the order of $10^3$ triangles) and use SDF mesh colliders generated from these meshes.
CRISP supports both custom SDFs and a torus-specialized TDSF representation; accordingly, we evaluate DSF--TDSF (peg as DSF, hole as TDSF) and SDF--SDF geometric configurations with both the CANAL and SubADMM solvers.
Since the clearances are extremely tight, we set the friction coefficient to 0.01 and apply a small angular perturbation of 0.5~\unit{\degree} without inducing jamming during insertion.

Since none of the compared simulators, including ours, are strictly intersection-free~\cite{li2020ipc}, we measure the average penetration depth throughout the insertion process as a quantitative metric of physical accuracy, as summarized in Table~\ref{tab:penetration}.
Overall, CRISP exhibits the lowest penetration across almost all tolerance levels.
The MuJoCo SDF--SDF baseline shows the largest penetration, which can be attributed to two main factors:
1) its SDF--SDF collision detection relies on nonconvex optimization, where insufficient initialization can miss contact points, particularly when the analytic SDFs exhibit discontinuous gradients as in the present setup; and
2) the Newton solver in MuJoCo employs numerical compliance to regularize contact constraints~\cite{todorov2014ccp}, inherently allowing penetration.
These results suggest that CRISP benefits from the DSF--TDSF representation, which provides more stable contact detection than SDF--SDF in this scenario, as well as from direct NCP handling even in the SDF--SDF configuration.
Isaac Sim, on the other hand, resolves penetration via positional constraint projection~\cite{muller2020xpbd}, yielding competitive metrics in this evaluation.
However, since its SDFs are internally generated from voxelized representations, contact accuracy is limited, and we observe failure cases under the tightest tolerance (i.e., 50~\unit{\micro\meter}).

\subsection{Bolt-Nut Assembly} \label{subsec:bolt_nut}
We next consider a bolt-nut assembly task, which involves intensive multi-contact interactions along helical threads, as shown in Fig.~\ref{fig:assembly}.
We primarily define the geometries using SDFs, based on analytic SDF formulations from the MuJoCo codebase~\cite{mujoco2025sdf}.
For Isaac sim, we construct meshes with identical geometry, which are also evaluated in CRISP as Mesh--SDF case.
We set the friction coefficient to zero to allow gravity-driven assembly and measure the average penetration depth in the same manner, as summarized in Table~\ref{tab:penetration}.

Consistent with the previous scenario, CRISP with the CANAL solver achieves the lowest penetration.
In contrast, due to the intensive contact inherent in threaded assembly, we observe that the SubADMM solver suffers from limited convergence due to its first-order nature, leading to larger penetration.
Unlike the peg insertion case, the SDF--SDF configuration in CRISP exhibits only a moderate advantage over MuJoCo in this scenario.
We attribute this reduced advantage to the use of numerical differentiation for computing SDF gradients required by SDF--SDF collision detection, which leads to inaccuracies in contact point and normal estimation.
When using the Mesh--SDF configuration, the mesh triangles provide effective guidance for initial contact points, enabling accurate collision detection and minimal penetration at moderate tolerances, while at the tightest tolerance the assembly fails due to limitations imposed by the mesh discretization.

\subsection{Top-Heavy Stacking} \label{subsec:stacking}
\begin{figure}[t]
  \centering
  \begin{minipage}{0.22\linewidth}
    \centering
    \includegraphics[width=\linewidth]{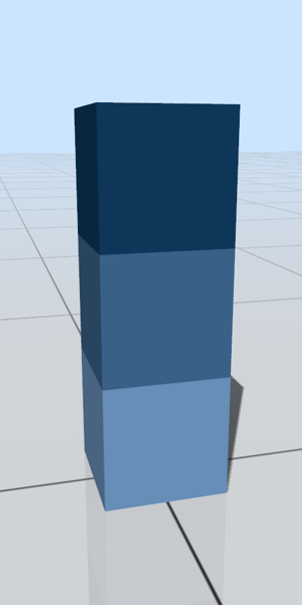}
  \end{minipage}\hfill
  \begin{minipage}{0.75\linewidth}
    \centering
    \includegraphics[width=\linewidth]{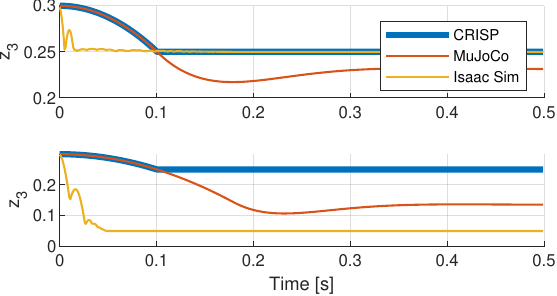}
  \end{minipage}
  \caption{
    \textbf{Top-heavy stacking scenario.}
    Left: Visualization of the scenario.
    Right: Height of the top block over time for mass ratios of 10 (top) and 100 (bottom).
  }
  \label{fig:stacking}
\end{figure}
In this scenario, we evaluate the robustness in handling ill-conditioned contact problems by stacking blocks with highly unbalanced mass ratios, where a heavier block is placed on top of lighter ones, all of which are cubic blocks with an edge length of 0.1~\unit{\meter}, as illustrated in Fig.~\ref{fig:stacking}.
We compare the height trajectories when the third block from the top is dropped slightly from above, while varying the mass ratio between adjacent blocks as 10 and 100.
As shown in Fig.~\ref{fig:stacking}, CRISP successfully simulates the stacking process without instability for both mass ratios.
In contrast, MuJoCo exhibits oscillatory behavior, particularly at a mass ratio of 100, which we attribute to regularization in its contact formulation that introduces artificial compliance~\cite{castro2023sap}.
Isaac Sim shows stable behavior in mass ratio of 10 with minor bouncing, but becomes unstable at a mass ratio of 100, likely due to the limited accuracy of its position-based contact solver~\cite{muller2020xpbd} when dealing with ill-conditioned contacts.

\subsection{Oblique Sliding} \label{subsec:sliding}
\begin{figure}[t]
  \centering
  \begin{minipage}{0.48\linewidth}
    \centering
    \includegraphics[width=\linewidth]{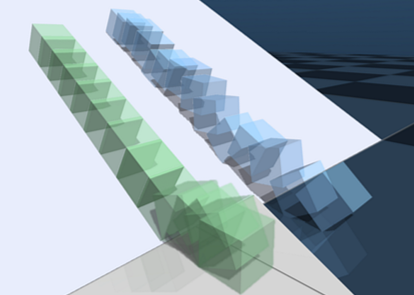}
  \end{minipage}\hfill
  \begin{minipage}{0.5\linewidth}
    \centering
    \includegraphics[width=\linewidth]{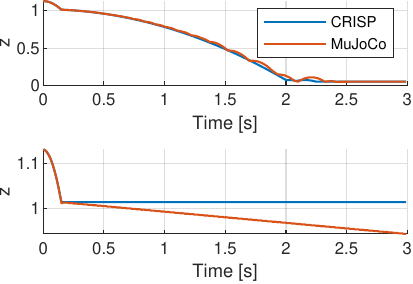}
  \end{minipage}
  \caption{
    \textbf{Oblique sliding scenario.}
    Left: Trajectories for the sliding case in CRISP (green) and MuJoCo (blue).
    Right: Height of the block over time for sliding case (top) and sticking case at the critical friction coefficient (bottom).
  }
  \label{fig:oblique}
\end{figure}
We further evaluate the accuracy of frictional contact handling through an oblique sliding test, where a block is placed on an inclined plane, as illustrated in Fig.~\ref{fig:oblique}.
The inclination angle is set to 30\unit{\degree}, with friction coefficients of $\tan(25\unit{\degree})$ to induce sliding and $\tan(30\unit{\degree})$ to achieve sticking.
While CRISP yields the expected sliding and sticking behaviors, MuJoCo exhibits deviations in both cases: in the sliding case, the CCP formulation induces a \textit{gliding} artifact~\cite{castro2023sap} that causes the block to lift off the inclined plane, and even in the sticking case, its contact regularization in tangential direction leads to creeping motion despite the presence of friction.
In contrast, Isaac Sim exhibits relatively stable behavior due to its position-level constraint projection; however, we observe that the simulation errors are inconsistent with respect to the initial conditions.

\subsection{Joint Constraints} \label{subsec:joint}
\begin{figure}[t]
  \centering
  \includegraphics[width=\linewidth]{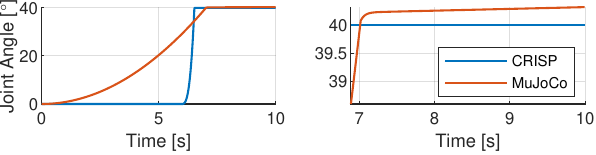}
  \caption{
    \textbf{Joint constraint scenario.}
    Joint angle trajectories over time (left) and a magnified view near the final time step (right).
  }
  \label{fig:joint}
\end{figure}
The last evaluation scenario focuses on the accurate enforcement of joint constraints.
We simulate a simple door handle mechanism consisting of a revolute joint with joint limits of $\pm40\unit{\degree}$ with friction, then apply a linearly increasing torque to the handle.
As shown in Fig.~\ref{fig:joint}, CRISP accurately enforces the joint limits and friction, with the handle coming to a stop just before reaching the limit.
In contrast, MuJoCo fails to completely suppress the applied torque through joint friction, and its soft enforcement of joint limits leads to observable limit violations.
We omit Isaac Sim from this comparison because an equivalent joint-friction setup was not available in our evaluation implementation.

\subsection{Demonstration Scenarios} \label{subsec:demo}
We now demonstrate the simulation capabilities of CRISP through robotic manipulation scenarios involving complex geometric configurations and rich multi-contact interactions.

\subsubsection{Gear Insertion and Driving}
\begin{figure}[t]
  \centering
  \includegraphics[width=\linewidth]{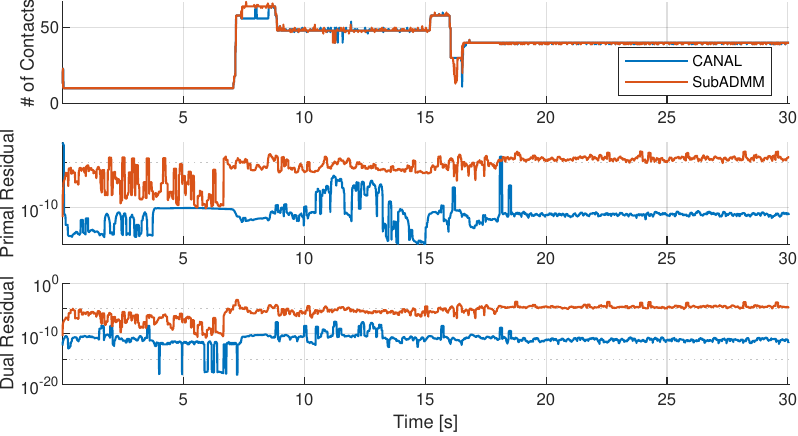}
  \caption{
    \textbf{Results of gear insertion and driving simulation.}
    Number of contacts and solver residuals as a function of time.
  }
  \label{fig:gear}
\end{figure}
We first consider a gear insertion and driving task, as illustrated in Fig.~\ref{fig:highlight}.
In this scenario, a robotic arm inserts a gear into a housing, after which the gear is driven by an adjacent pre-assembled gear through sustained frictional contact.
From a simulation perspective, this scenario is challenging due to the presence of simultaneous, contact-rich interactions among multiple object pairs during insertion and driving.
The gear and housing are modeled using analytic SDFs, and the simulation capabilities are evaluated using both the CANAL and SubADMM solvers.
As shown in Fig.~\ref{fig:gear}, the evolution of contact constraints and solver residuals over time indicates stable and robust simulation behavior, with the primal residual $\|Jv - z\|$ measuring constraint violation and the dual residual $\|Av - b - J^T\lambda\|$ quantifying deviation from the system dynamics.

\subsubsection{Robotic Bolt-Nut Assembly}
As a second demonstration, we implement a robotic bolt-nut assembly task, shown in Fig.~\ref{fig:highlight}.
In this setup, a robotic arm equipped with a continuous rotary gripper with the nut attached approaches a bolt fixed to the ground, and performs the assembly by aligning and rotating the gripper along the thread axis.
Following the evaluation results in Sec.~\ref{subsec:bolt_nut}, we use analytic SDF and mesh to represent the bolt and nut, respectively, to accurately capture the contact interactions along the threads.
This scenario is characterized by dense and stiff contact interactions, which impose tightly coupled contact constraints under limited effective degrees of freedom, yet are handled robustly within CRISP to achieve stable and successful simulation of the assembly process.

\section{Conclusion} \label{sec:conclusion}
In this paper, we present CRISP, a new simulation platform to achieve high physical fidelity in contact-rich robotic scenarios.
Our simulator supports diverse geometric representations and accurate optimization-based collision detection, and combines multi-contact constraint modeling with robust AL-based constraint solvers.
Extensive evaluations demonstrates that CRISP reliably handles challenging contact scenarios and achieves improved accuracy and stability compared to state-of-the-art simulators.

By consistently resolving tightly coupled multi-contact interactions without unphysical compromises, CRISP is particularly well suited for contact-intensive and tight-tolerance tasks such as bolting, insertion, and complex assembly, which are common in industrial environments.
These capabilities make CRISP a practical simulation platform for both control and learning-based robotic systems.

Since CRISP is designed as a flexible and extensible platform, we plan to further enhance its capabilities along several directions.
Future work will extend supported collision pairs, improve solver performance through preconditioning and parallelization, incorporate additional physical phenomena such as soft-body dynamics, and further enhance scalability across larger contact-rich systems.

\bibliographystyle{plainnat}
\bibliography{references}

\begin{table*}[t]
  \centering
  \setlength{\extrarowheight}{2pt}
  \begin{tabular}{|l||cccccccccc|}
    \hhline{~|----------|}
    \multicolumn{1}{c|}{}
    & Plane & Sphere & Capsule & Box & Cylinder & Convex & Mesh & SDF & DSF & TDSF \\
    \hhline{-::==========}

    Plane
    & \multicolumn{1}{c|}{--}
    & \multicolumn{4}{c|}{\cellcolor{lvlPerfect!20}} 
    & \multicolumn{1}{c|}{\cellcolor{lvlPerfect!20}SF}
    & \multicolumn{1}{c|}{\cellcolor{lvlMid!20}} 
    & \multicolumn{1}{c|}{\cellcolor{lvlMid!20}GD}
    & \multicolumn{2}{c|}{\cellcolor{lvlPerfect!20}SF} \\
    \hhline{|~||-|~~~~|-|~|-|--|}

    Sphere
    & \multicolumn{1}{c|}{}
    & \multicolumn{4}{c|}{\cellcolor{lvlPerfect!20}Analytic}
    & \multicolumn{1}{c|}{\cellcolor{lvlHigh!20}} 
    & \multicolumn{1}{c|}{\cellcolor{lvlMid!20}\multirow{-2}{*}{BF}}
    & \multicolumn{1}{c|}{\cellcolor{lvlPerfect!20}Direct}
    & \multicolumn{2}{c|}{\cellcolor{lvlHigh!20}RTR} \\
    \hhline{|~||~|-|~~|-|~|-|-|--|}

    Capsule
    & \multicolumn{2}{c|}{}
    & \multicolumn{2}{c|}{\cellcolor{lvlPerfect!20}} 
    & \multicolumn{2}{c|}{\cellcolor{lvlHigh!20}} 
    & \multicolumn{1}{c|}{\cellcolor{lvlHigh!20}} 
    & \multicolumn{1}{c|}{\cellcolor{lvlMid!20}Search}
    & \multicolumn{1}{c|}{\cellcolor{lvlHigh!20}} 
    & \multicolumn{1}{c|}{\multirow{6}{*}{$\times$}} \\
    \hhline{|~||~~|-|~|~~|~|-|~~|}

    Box
    & \multicolumn{3}{c|}{}
    & \multicolumn{1}{c|}{\cellcolor{lvlPerfect!20}} 
    & \multicolumn{2}{c|}{\cellcolor{lvlHigh!20}MPR}
    & \multicolumn{1}{c|}{\cellcolor{lvlHigh!20}FW}
    & \multicolumn{1}{c|}{\cellcolor{lvlMid!20}} 
    & \multicolumn{1}{c|}{\cellcolor{lvlHigh!20}} 
    & \\
    \hhline{|~||~~~|-|~~|~|~|~|~|}

    Cylinder
    & \multicolumn{4}{c|}{}
    & \multicolumn{2}{c|}{\cellcolor{lvlHigh!20}} 
    & \multicolumn{1}{c|}{\cellcolor{lvlHigh!20}} 
    & \multicolumn{1}{c|}{\cellcolor{lvlMid!20}\multirow{-2}{*}{GD}}
    & \multicolumn{1}{c|}{\cellcolor{lvlHigh!20}} 
    & \\
    \hhline{|~||~~~~|-|~|--|~|~|}

    Convex
    & \multicolumn{5}{c|}{}
    & \multicolumn{1}{c|}{\cellcolor{lvlHigh!20}} 
    & \multicolumn{1}{c|}{\cellcolor{lvlLow!20}} 
    & \multicolumn{1}{c|}{\cellcolor{lvlHigh!20}} 
    & \multicolumn{1}{c|}{\cellcolor{lvlHigh!20}\multirow{-4}{*}{MPR}}
    & \\
    \hhline{|~||~~~~~|-|~|~|-|~|}

    Mesh
    & \multicolumn{6}{c|}{}
    & \multicolumn{1}{c|}{\cellcolor{lvlLow!20}\multirow{-2}{*}{MPR}}
    & \multicolumn{1}{c|}{\cellcolor{lvlHigh!20}\multirow{-2}{*}{FW}}
    & \multicolumn{1}{c|}{\cellcolor{lvlLow!20}MPR}
    & \\
    \hhline{|~||~~~~~~|-|-|-|~|}

    SDF
    & \multicolumn{7}{c|}{}
    & \multicolumn{1}{c|}{\cellcolor{lvlMid!20}GD}
    & \multicolumn{1}{c|}{\cellcolor{lvlMid!20}RTR}
    & \\
    \hhline{|~||~~~~~~~|-|--|}

    DSF
    & \multicolumn{8}{c|}{}
    & \multicolumn{2}{c|}{\cellcolor{lvlHigh!20}RTR} \\
    \hhline{|~||~~~~~~~~|--|}

    TDSF
    & \multicolumn{9}{c|}{}
    & \multicolumn{1}{c|}{\cellcolor{lvlMid!20}RTR} \\
    \hhline{|-||----------|}
  \end{tabular}
  \caption{
    \textbf{Collision function table in CRISP.}
    Each cell indicates the collision detection algorithm used for the corresponding geometry pair (--: cannot be defined, $\times$: not supported).
    Cell colors indicate a qualitative assessment of algorithmic performance (green: highly reliable, blue: recommended, yellow: neutral, red: use with caution).
  }
  \label{tab:collision_table}
\end{table*}

\newpage
\appendix
\subsection{Classes of DSF} \label{app:dsf}
CRISP currently supports two classes of DSFs, as illustrated in Fig.~\ref{fig:dsf}.
The first is the vertex-based DSF originally proposed in \cite{lee2023dsf}, which approximates a convex hull with vertices $\{v_1,\cdots,v_n\}\subset\mathbb{R}^3$ and defined by
\begin{equation*}
  h(x) = \left(\sum_{i=1}^n\{\max(v_i\cdot x,0)\}^p\right)^\frac{1}{p}
\end{equation*}
where $p>2$ denotes the sharpness of the smoothed shape.

The second is the axis-based DSF proposed in \cite{an2024dsf}, which correponds to a smooth convex shape generated by rotating a convex hull defined on $xz$-plane with vertices $\{v_1,\cdots,v_n\}\subset\mathbb{R}^2$ symmetric to $z$-axis and defined by
\begin{equation*}
  h(x) = \left(\sum_{i=1}^n\{\max(v_i\cdot x_{rz},0)\}^p\right)^\frac{1}{p}
\end{equation*}
where $x_{rz}:=(\sqrt{x_1^2+x_2^2+\epsilon x_3^2},x_3)$ and $\epsilon>0$ determines the flatness at the poles.

\subsection{Collision Function Table} \label{app:collision_table}
For each geometry pair, CRISP dispatches the most suitable collision detection algorithm based on the characteristics of the involved geometries, as summarized in Table~\ref{tab:collision_table}.

\subsubsection{Convex Collisions}
Collision detection between convex geometries can be formulated as a convex optimization problem, enabling efficient algorithms with strong convergence properties~\cite{gilbert2002gjk}.
While a single contact generally suffices for strictly convex pairs, weakly convex shapes such as boxes and convex hulls may require multiple contacts to represent edge or face contact robustly~\cite{drumwright2019contact}.
CRISP uses specialized analytic routines for primitive pairs, including the separating axis theorem for box--box collisions~\cite{boyd2004convex}, and support-function (SF) queries when a plane is paired with a support-mapped geometry.
For general convex pairs without a specialized routine, CRISP employs Minkowski portal refinement (MPR)~\cite{snethen2008mpr} with rotational perturbations to generate multiple contacts when needed, while DSF--DSF pairs use the dedicated Riemannian trust-region (RTR) method~\cite{an2024dsf}.

\subsubsection{Non-Convex Collisions}
Collision detection involving non-convex geometries is more challenging because multiple local solutions may exist, motivating pair-specific algorithms that exploit the available geometric structure.
These include brute-force sampling (BF) for simple mesh pairs, Frank--Wolfe (FW) for Mesh--SDF, gradient descent (GD) for SDF--SDF, and RTR optimization for SDF--DSF and selected TDSF pairs; see Sec.~\ref{sec:collision_detection} and Table~\ref{tab:collision_table}.
For geometry pairs without a dedicated non-convex routine, a convex-hull fallback may be used where applicable, while the remaining combinations are not currently supported.

\subsection{Algorithmic Details on Constraint Solving} \label{app:constraint_solving}
The constraints supported in CRISP are categorized as follows:

\begin{itemize}
  \item \textbf{Limit}:
  For joints defined with upper and lower bounds, the following hard inequality constraints are imposed:
  \begin{equation*}
    0 \leq \lambda_i \perp J_iv + e_i \geq 0
  \end{equation*}
  where $e_i\in\mathbb{R}$ and $J_i\in\mathbb{R}^{1\times n_v}$ are the error and Jacobian for joint limit constraint, and $\perp$ denotes complementarity.

  \item \textbf{Friction}:
  For joints with friction parameters, the following dry friction constraint is imposed:
  \begin{equation*}
    \begin{cases}
      |\lambda_i| \leq b_i        & \text{if } J_iv = 0 \\
      \lambda_i = -b_iJ_iv/|J_iv| & \text{if } J_iv \neq 0
    \end{cases}
  \end{equation*}
  where $J_i\in\mathbb{R}^{1\times n_v}$ is the Jacobian for joint friction constraint and $b_i>0$ is the maximum friction impulse.

  \item \textbf{Contact}:
  The contact condition is the most challenging constraint as it is characterized by a complex nonlinear complementarity relation.
  We employ the following extended Signorini-Coulomb condition with compliance:
  \begin{equation} \label{eq:contact_constraint}
    \mathcal{C}_i \ni
    \lambda_i \perp J_iv + e_i + p_i + R_i\lambda_i
    \in \mathcal{C}_i^\ast
  \end{equation}
  where $J_i=[J_{i,t}; J_{i,n}]\in\mathbb{R}^{3\times n_v}$ and $e_i=[0; 0; e_{i,n}]\in\mathbb{R}^3$ are the contact Jacobian and error consisting of tangential and normal components, $p_i=[0; 0; \mu_i\|J_{i,t}v\|]$ is the De Saxc\'e correction term, $\mathcal{C}_i$ and $\mathcal{C}_i^\ast$ are the friction cone and its dual with coefficient $\mu_i>0$, and $R_i=\mathrm{diag}(0,0,c_i)$ introduces compliance $c_i\geq0$, with $c_i=0$ reducing to the rigid case.
\end{itemize}

For the AL defined by \eqref{eq:augmented_lagrangian}, the AL iteration step consists of solving the surrogate problem for optimality conditions~\eqref{eq:augmented_lagrangian_v},~\eqref{eq:augmented_lagrangian_z} and updating the multiplier~\eqref{eq:augmented_lagrangian_u} as follows:
\begin{align}
  \label{eq:augmented_lagrangian_v}
  &(A + \beta J^TJ)v = b + J^T(\beta z - u) \\
  \label{eq:augmented_lagrangian_z}
  &\beta z = \beta Jv + u + \lambda, \quad
  \text{s.t. } (z,\lambda)\in\mathcal{S}_c \\
  \label{eq:augmented_lagrangian_u}
  &u \leftarrow u + \beta(Jv - z)
\end{align}
We note that the closed-form solution to \eqref{eq:augmented_lagrangian_z} can be derived via the following projection operations for each constraint type:
\begin{equation} \label{eq:projection}
  \lambda_i = \Pi_i(-\beta J_iv - u_i - \beta e_i)
\end{equation}
For instance, regarding the contact constraint in \eqref{eq:contact_constraint}, the projection $\Pi_i$ corresponds to the projection onto the friction cone $\mathcal{C}_i$ while preserving the non-negative normal component:
\begin{align*}
  \lambda_{i,n} &=
    \begin{cases}
      \Pi_{\geq0}(-\beta J_{i,n}v - u_{i,n} - \beta e_{i,n})         & \text{if } c_i = 0 \\
      \Pi_{\geq0}(-c_i^{-1}(J_{i,n}v + \beta^{-1}u_{i,n} + e_{i,n})) & \text{if } c_i > 0 \\
    \end{cases} \\
  \lambda_{i,t} &= \Pi_{\|\cdot\|\leq\mu_i\lambda_{i,n}}(-J_{i,t}v - \beta^{-1}u_{i,t})
\end{align*}
where $\Pi_{\geq0}$ and $\Pi_{\|\cdot\|\leq r}$ denote the projection onto the non-negative reals and the ball of radius $r>0$, respectively.
By substituting \eqref{eq:projection} into \eqref{eq:augmented_lagrangian_v}, we obtain the semismooth equation~\eqref{eq:canal_residual} for the surrogate problem which is addressed in CANAL solver.

\subsection{Evaluation Setups} \label{app:evaluation_setup}
This section describes the experimental configurations for each scenario.
To ensure a fair comparison, we configure the geometric and physical settings to be as identical as possible across all simulators.
Unless otherwise specified, the simulation parameters were kept at their default values.

\subsubsection{Peg Insertion}
In this scenario, we evaluate the simulators by dropping the peg from a position directly above the hole with a slight tilt of 0.5$^\circ$.
To ensure geometric equivalence, both MuJoCo and CRISP utilize analytic SDFs for the peg and the hole, while Isaac Sim uses high-resolution meshes with the voxel SDF resolution set to its maximum to minimize discretization errors.
When CRISP employs DSF and TDSF geometries, the sharpness parameter is set to $p=500$ to ensure that any geometric deviations from surface smoothing remain negligible, thereby preserving the intended sharp features.
We use a time step of $\Delta t=2~\unit{\milli\second}$ across all engines.
To quantify the penetration depth, we consider a reduced 2D planar representation and calculate the maximum penetration depth for each time step during the insertion process and average these values over time.

\subsubsection{Bolt-Nut Assembly}
For this scenario, we utilize analytic SDFs provided by the MuJoCo codebase~\cite{mujoco2025sdf}, where the tolerance is controlled by adjusting the radius parameter of the bolt.
Specifically, for evaluations in Isaac Sim and CRISP (with Mesh--SDF collisions), we generate high-resolution meshes that precisely match the thread geometry defined by the analytic SDFs.
In Isaac Sim, we further set the voxel SDF resolution to its maximum to minimize discretization errors.
Since the complex geometries at the thread entrance and the connection to the bolt head cannot be perfectly matched between meshes and analytic SDFs, we collect data only during the middle segment of the engagement, where the nut is fully engaged with the threads.
A time step of $\Delta t=10~\unit{\milli\second}$ is used for this scenario.
Unlike the previous case where an explicit measurement of penetration depth is possible, the complex helical geometry of the threads precludes such a direct calculation; we therefore record the maximum penetration among all contact points at each time step and calculate their average for comparison.

\subsubsection{Top-Heavy Stacking}
In this scenario, we evaluate the stacking stability by varying the mass ratio of the top and bottom blocks to 10 and 100, while keeping the middle block at 1~\unit{\kilogram}.
To ensure the stack remains stable without slipping, the friction coefficient is set to 5.0, and a time step of $\Delta t=1~\unit{\milli\second}$ is used.
We note that while MuJoCo exhibits noticeable penetration in this setup, such effects can be reduced by tuning \texttt{solimp}, but finite penetration may remain under MuJoCo's soft-contact formulation.

We observe that since this configuration is underdetermined in terms of contact forces, CANAL and SubADMM exhibit different force distribution characteristics: while SubADMM tends to distribute contact forces across the contact surface, CANAL shows a tendency to concentrate forces at specific contact points.
Despite these different behaviors in force distribution, both methods successfully maintain a stable stack under all tested mass ratios.

\subsubsection{Oblique Sliding}
In the oblique sliding scenario, we evaluate the accuracy for frictional behavior by dropping a 1 kg block, rotated to align with the slope's inclination, from a height of 0.1 m above the plane.
We use a time step of $\Delta t=10~\unit{\milli\second}$ for this evaluation.
Similar to the previous scenarios, we note that gliding or creeping in MuJoCo can be mitigated by tuning parameters such as \texttt{impratio} or \texttt{solimp}, but may persist under the compliant convex contact formulation used by MuJoCo.

\subsubsection{Joint Constraints}
In this scenario, we model a door handle mechanism to evaluate the precision of joint constraints.
The joint is configured with a friction torque of 0.06~\unit{\newton\meter} and limits set at $\pm40\unit{\degree}$.
To isolate the joint's response to external inputs, we disable gravity and attach a force-controlled actuator, with the input torque increasing linearly over time.
A time step of $\Delta t=10~\unit{\milli\second}$ is applied across both simulators.

\subsubsection{Demonstration Scenarios}
For both demonstration scenarios, we utilize a time step of $\Delta t=10~\unit{\milli\second}$.
To control the robotic manipulator, we employ a low-level PD controller for each joint where the desired joint angles are determined by interpolating between a set of predefined keyframes.

\end{document}